%% file: v4.tex
\documentclass[letterpaper, 10 pt, journal, twoside]{IEEEtran}  

\IEEEoverridecommandlockouts                              

\usepackage{graphics} 
\usepackage{amsmath} 
\usepackage{amssymb}  

\newcommand*{\V}{\mathbf}

\usepackage{mathtools}

\usepackage{changepage} 
\usepackage{pstool}
\usepackage{subcaption}
\usepackage{caption}
\usepackage{array}
\usepackage[dvipsnames]{xcolor}
\usepackage{hyperref} 
\usepackage[capitalize]{cleveref}
\usepackage{algpseudocode}
\usepackage[]{algorithm2e}
\usepackage{multirow}
\usepackage{units}
\definecolor{magenta1}{rgb}{0.761, 0.428, 0.627}
\usepackage{epstopdf}
\usepackage{booktabs} 

\title{
  Online Learning of Neural Surface Light Fields\\ alongside Real-time Incremental 3D Reconstruction
}
\author{Yijun Yuan and Andreas N\"uchter
  \thanks{Manuscript received: January, 24, 2023; Revised March, 29, 2023; Accepted April, 28, 2023.}
  \thanks{This paper was recommended for publication by Editor Civera Javier upon evaluation of the Associate Editor and Reviewers' comments.}
  \thanks{All authors are with Informatics XVII : Robotics, Julius-Maximilians-Universit\"at W\"urzburg, Germany.
    {\tt\footnotesize \{yijun.yuan|andreas. nuechter\}@uni-wuerzburg.de }}%
}

\begin{document}

\maketitle
\markboth{IEEE Robotics and Automation Letters. Preprint Version. Accepted April, 2023}
{Yuan \MakeLowercase{\textit{et al.}}: Neural Surface Light Fields alongside Reconstruction}

\begin{abstract}

Immersive novel view generation is an important technology in the field of graphics and has recently also received attention for operator-based human-robot interaction. 
However, the involved training is time-consuming, and thus the current test scope is majorly on object capturing.
This limits the usage of related models in the robotics community for 3D reconstruction since robots (1) usually only capture a very small range of view directions to surfaces that cause arbitrary predictions on unseen, novel direction, (2) requires real-time algorithms, and (3) work with growing scenes, e.g., in robotic exploration.
The paper proposes a novel Neural Surface Light Fields model that copes with the small range of view directions while producing a good result in unseen directions.
Exploiting recent encoding techniques, the training of our model is highly efficient.

In addition, we design Multiple Asynchronous Neural Agents (MANA), a universal framework to learn each small region in parallel for large-scale growing scenes.
Our model learns online the Neural Surface Light Fields (NSLF) aside from real-time 3D reconstruction with a sequential data stream as the shared input.
In addition to online training, our model also provides real-time rendering after completing the data stream for visualization.
We implement experiments using well-known RGBD indoor datasets, showing the high flexibility to embed our model into real-time 3D reconstruction and demonstrating high-fidelity view synthesis for these scenes. The code is available on github\footnote{\url{https://jarrome.github.io/NSLF-OL}}.
\end{abstract}
\vspace{-.3cm}
\input{tex/intro}

\input{tex/related}

\input{tex/method}

\input{tex/exp}

\vspace{-.2cm}
\section{Conclusion}
In this paper, we have proposed an online learning method for neural surface light fields during real-time incremental 3D reconstruction on large scenes.

We have proposed a novel Neural Surface Light Fields model to address the challenge that in a SLAM and reconstruction scenario the captured surface directions are very limited, the learned model easily produces arbitrary predictions from unseen directions.

For online learning in growing scenes where we do not pre-know the boundaries in advance, we have designed Multiple Asynchronous Neural Agents to work alongside real-time incremental 3D reconstruction.

The performance of the proposed method has been demonstrated in our experiments.
Our implementation achieves real-time learning of Neural Surface Light Fields alongside real-time incremental reconstruction.

{\small
	\bibliographystyle{IEEEtran.bst}
	\bibliography{ref}
}

\end{document}

%% file: tex/intro.tex
\section{Introduction}

\IEEEPARstart{I}{n} robotics, mapping, and 3D reconstruction have long been of great interest and have been studied for decades.
Initially, researchers focused on 3D point clouds and voxel grids, and this later shifted towards using Signed Distance Functions (SDF).
%
An SDF, is the starting point of many of the following state-of-the-art papers.
It has been well-developed from Point-to-SDF~\cite{bylow2013real} to SDF-to-SDF~\cite{slavcheva2016sdf,yuan2022indirect}, from explicit voxel field~\cite{niessner2013real} to neural implicit representation~\cite{huang2021di,yuan2022algorithm}. 
These methods have achieved high-quality 3D reconstructions in real-time while maintaining high regression functionality of the neural models.

Aside from the reconstruction of geometries, neural rendering for colors is also a hot topic, however more in the field of computer graphics~\cite{tewari2021advances}.
Neural rendering focuses on the synthesis of novel views from 3D models. 
The goal of this topic is to enhance the immersive experience for users. In the context of robotics, for instance, it is crucial for human-robot interaction or situation awareness of an operator of a telerobot. 
In 2020, the Neural Radiance Field (NeRF)~\cite{mildenhall2020nerf} was introduced as an innovative approach to high-resolution rendering for view synthesis, activating the trend of neural rendering.
It learns a neural radiance field that produces both an occupancy field and a light field via differentiable rendering for custom immersive view synthesis, and it has extended the testing scope of 3D reconstruction to large scene-level
~\cite{tancik2022block,zhang2022nerfusion}.
The drawback is the extremely high training time, and it is therefore unsuitable for real-time 3D reconstruction.
Nevertheless, it still affects the field of 3D reconstruction.
Recent works on 3D reconstructions, i.e., iMAP~\cite{sucar2021imap} and NICE-SLAM~\cite{zhu2022nice}, build on the differentiable rendering from NeRF to approach an online reconstruction with color.
However,
their visualization of color appears blurred.
\begin{figure}[t]
	\centering
	\psfragfig[width=1\linewidth]{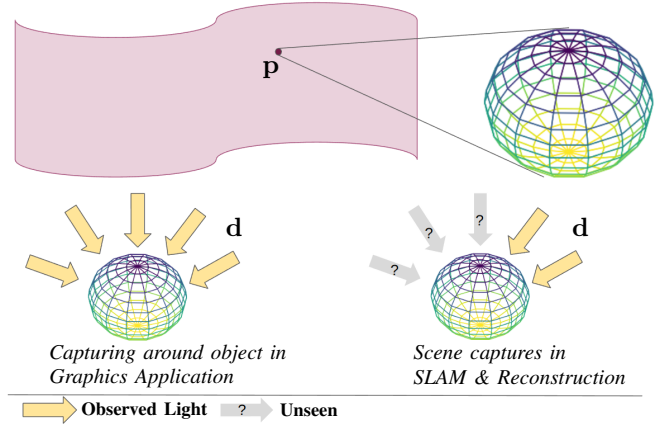}{
		\psfrag{p}{$\V p$}
		\psfrag{d}{$\V d$}
		\psfrag{T1}{\textit{\footnotesize Capturing around object in}}
		\psfrag{Tn1}{\textit {\footnotesize Graphics Application}}
		\psfrag{T2}{\textit{\footnotesize Scene captures in}}
		\psfrag{Tn2}{\textit{\footnotesize SLAM \& Reconstruction}}
		\psfrag{T3}{\scriptsize\textbf{Observed Light}}
		\psfrag{T4}{\scriptsize\textbf{Unseen}}
	}
	\caption{Light difference between object capturing for graphics applications and scene capturing in robotic SLAM \& reconstruction. At 3D point $\V p$, lights are cast from direction $\V d$. The \textbf{sphere} shows the $\mathbf S^2$ space for the direction vectors $\V d$ that are partially covered by light rays.}
	\label{fig:limited_light}
	\vspace{-.5cm}
\end{figure}
It is worth mentioning that shape reconstruction has already been extensively studied and has already achieved high-quality 3D reconstruction performance. Aside from the high quality, the robotics community values online capable reconstructions.

However, NeRFs in the graphics community concentrate \emph{more} on (1) object capturing instead of capturing large scene surfaces (captures contain dense view-directions to surface), (2) rendering high-quality images while caring less on depth and surface accuracies (shape-radiance ambiguity)~\cite{zhang2020nerf++}, and (3) more on rendering (testing) speed instead of online training.
Therefore, we explore the high potential to use surface reconstruction research as the basis of this topic to meet the interest of the robotics community to have a universal way to simulate real environments.

We utilize a real-time reconstruction model to provide surfaces and simplify the problem to online learn the light field on surfaces \textbf{aside from} real-time reconstruction as given in \cref{fig:workflow}. Where our model relies on data from reconstruction without affecting it. During the online learning phase, we employ a colored point cloud. Subsequently, in the inference phase, we apply ray-rasterizing on the reconstruction mesh to obtain the point cloud for color prediction.

The Surface Light Fields (SLF) has been proposed by Wood et al.~\cite{wood2000surface} to model the surface reflectance.
For a surface point $\V p$, the radiance of a reflection ray (with direction $\V d_o$) is computed from an accumulation of incident rays (with direction $\V d_i$ for ray $i$).
However, due to the inefficient dense sampling requirement of SLF, more work has focused on modeling compression~\cite{miandji2013learning}. 
Recently, Chen et al.~\cite{chen2018deep} and Yu et al.~\cite{yu2022anisotropic} have introduced Deep Surface Light Fields (DSLF), which utilize a neural network to replace the previous handcrafted formulation.
Still, these approaches have an extremely slow training speed. 
Similarly, these techniques are mainly concentrated on capturing a 360-degree view of the object.
Large-scale scenes have not been considered.

Unlike the graphics communities, in robotics, SLAM, and 3D reconstruction \emph{(1) capture scene frames that contain only a limited range of surface view-direction} as depicted \cref{fig:limited_light}, apply for \emph{(2) growing large-scale scenes without prior knowledge of their size} and require \emph{(3) real-time processing speed}.
These three points are the primary concerns for using Neural Surface Light Fields (NSLF) in robotics.
Thus, this paper aims to address the limited view-direction challenge during testing, where most unseen directions will cause arbitrary results.
To cope with this problem, we introduce Spherical Harmonics for \emph{decoding}, as depicted in~\cref{fig:NSLF}.
Furthermore, relying on recent advances in graphics, i.e., the Multi-resolution Hash-encoding, we train grid-latent as an encoder for (3) real-time online learning.
We further introduce Multiple Asynchronous Neural Agents (MANA) to handle (2), i.e., we handle large scenes without pre-knowing the size. 

We take inspiration from~\cite{reiser2021kilonerf} which claims that multiple local MLPs converge faster. 
However, \cite{reiser2021kilonerf} has to train all local MLPs with one optimizer because NeRF's differentiable rendering accumulates data from different MLPs. 
While Surface Lighting Field training does not depend on ray integration in NeRF. Therefore, each region has its model and optimizer to learn individually.

To deal with scalable data when data is distributed, we dynamically allocate neural models and optimizers for new regions and run them in a new thread independently as depicted in \cref{fig:workflow}.
Every region has a neural model and an optimizer running independently without synchronizing with others.

The contributions of this paper are as follows:
\begin{itemize}
\item
  Proposing a novel Neural Surface Light Fields model to address the issue of arbitrary prediction on unseen direction causing from the small range of view-direction to surface in SLAM \& 3D reconstruction captures,
\item
  Proposing the first framework (MANA) for online-learning neural surface light fields on growing large-scale scenes,
\item
  Implementing MANA aside from real-time reconstruction for experiments.
  Both online learning and real-time testing are supported.
  Agents can be distributed to multiple GPUs.
\end{itemize}

%% file: tex/related.tex
\section{Related Works}

Our work is mainly related to NSLF.
Here, we also review the hotter sister topic, NeRF because its development process highly inspires and directs us to use NSLF online for the large scenes. 

\subsection{Neural Radiance Fields (NeRF)}

Mildenhall et al.~\cite{mildenhall2020nerf} propose to represent a scene as a continuous mapping from five-dimensional coordinates $(x,y,z,\theta,\phi)$ to volume density and view-dependent RGB color $(\sigma RGB)$, which are so-called NeRF representations.
The training of NeRF relies on the use of differentiable rendering along rays.
NeRF represents a significant advancement in graphics due to its immersive view-synthesis performance.
However, NeRF does have some limitations. Firstly, the speed for training and rendering is extremely slow. Secondly, the NeRF model suffers from shape-radiance ambiguity.

\subsubsection{Speed-aware NeRF}

Although NeRFs show a very high quality of view synthesis, their computational cost for training and testing is extremely high.
Therefore, lots of recent work shows interest in the speed issue. 
Tewari et al.~\cite{tewari2021advances} summarize and show many works exploring speedups using pruning~\cite{liu2020neural}, sampling~\cite{neff2021donerf}, fast integration~\cite{lindell2021autoint}, data structure~\cite{yu2021plenoctrees,hedman2021baking,muller2022instant}, and so on~\cite{reiser2021kilonerf,chen2022tensorf,sitzmann2021light}.
Many different data structures are also used to speed up the rendering and the training process benefits.
A recently highlighted work from NVIDIA, Instant Neural Graphics Primitives, uses a multi-resolution hash encoding to realize fast NeRF training in a few seconds~\cite{muller2022instant}. 
It provides an encoder with very fast convergence.
We will also employ this encoder in our proposed model.  
In the most related scope, KiloNeRF, which uses thousands of tiny MLPs to imitate standard NeRF, shows a factor of 2000 speedup in rendering~\cite{reiser2021kilonerf}.
Real-time rendering already meets the needs of the field of computer graphics (CG).

However, from a robotics perspective, online algorithms such as 3D reconstruction and SLAM are always preferred.
In such an environment, online training makes the task of light fields much more challenging.

\subsubsection{Depth Aided NeRF}

Another problem with NeRF is shape-radiance ambiguity.  NeRF++~\cite{zhang2020nerf++} points out that given an incorrect shape, there exists a family of radiance fields that perfectly explains the training images, but generalizes poorly to novel views. 
Although a well-trained NeRF does not show this phenomenon, imperfect depth prediction is common but is neglected since the application targets color prediction.
However, this also leads to another research direction, a group of works uses depth as an input to reduce the complexity of geometry modeling. 
This group of works embed Structure-from-motion (SfM)~\cite{deng2021depth,xu2022point}, Multi-view Stereo (MVS)~\cite{chen2021mvsnerf,zhang2021learning} or directly input RGBD data~\cite{azinovic2021neural}. 

The above algorithms show a trend that the facilitation of geometry learning well-reduces the complexity of NeRF while improving the high quality.
Moreover, in the robotics community, recent mature real-time 3D reconstructions already provide high-quality online shape reconstruction~\cite{huang2021di}. 

Therefore, we consider solving the geometric and color regression separately. Where the geometry is estimated using a mature online 3D reconstruction model. Meanwhile, an NSLF is independently trained for surface colors.
Without ray sampling, our model uses only surface points during training and inference.
This breaks the constraint of NeRF's rendering loss and allows asynchronous training and inference for each intelligent agent.

\subsection{Neural Surface Light Fields (NSLF)}

NSLF learns a mapping function $f_{NSLF}:\V X\times \V S^3 \rightarrow \V C$ where $\V X \subseteq \mathbb R^3$ denotes the point set on the surface, $\V S^3$ is the unit 3-sphere, $\V C$ denotes the RGB color space.

With the high regression capability of deep learning, Chen et al.~\cite{chen2018deep} propose Deep Surface Light Fields (DSLF) which uses MLPs to replace the accumulation function.
This parametric modeling greatly reduces the computational and spatial burden.
And thus provides more efficient rendering.
Yu et al.~\cite{yu2022anisotropic} have introduced Anisotropic Fourier Features Mapping (AFFM) to encode points.
This increases convergence speed and rendering quality.

Similar to the NeRF design, the NSLFs also follow the encoding-decoding pattern.
For a given point, its position vector is encoded as $\V F_{\V p}=\phi_{p}(\V p)$.
In the other branch, the direction vector is also encoded with a non-parameterized encoder, spherical harmonics, or frequency encoding: $\V F_{\V d} = \varphi_{sh/freq}(\V d)$.
Shallow MLPs are then applied to concatenated features (position encoding and direction) to predict color: $\V c = \phi_{dec}(\V F_{\V p}, \V F_{\V d})$. 

However, similar to NeRF the NSLF has not been used in large-scale incremental 3D reconstruction.
Thus, in the following, we fill this gap and provide an online learned neural surface light fields function for large scales.

%% file: tex/method.tex
\section{Methodology}
In the following, we introduce our proposed NSLF model, that addresses the problem of limited view directions toward the surfaces.
Then we further design a Multiple Asynchronous Neural Agents (MANA) framework that learns the NSLF alongside the reconstruction.
The geometry regression that provides the surface relies on the existing incremental 3D reconstruction models. 
\vspace{-.3cm}
\subsection{NSLF Design}
In the use of capturing of 360 degrees of an object, for a certain surface point, the trained view directions are densely sampled over a large range and thus, the novel view inference is mostly on the touched view directions.
However, in large-scale 3D SLAM and reconstruction, this feature of data acquisition is not guaranteed. 
Therefore, training on a small range of view directions will lead to arbitrary prediction on unseen views.

\begin{figure}[t]
	\centering
	\psfragfig[width=1\linewidth]{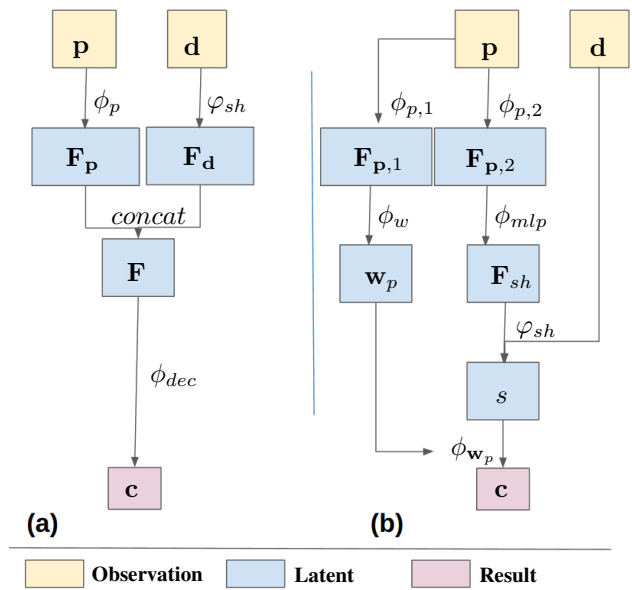}{
		\psfrag{p1}{$\V p$}
		\psfrag{d1}{$\V d$}
		\psfrag{p2}{$\V p$}
		\psfrag{d2}{$\V d$}
		\psfrag{f1}{$\phi_p$}
		\psfrag{f2}{$\varphi_{sh}$}
		\psfrag{f3}{$\phi_{p,1}$}
		\psfrag{f4}{$\phi_{p,2}$}
		\psfrag{f5}{$concat$}
		\psfrag{f6}{$\phi_{dec}$}
		\psfrag{f7}{$\phi_{w}$}
		\psfrag{f8}{$\phi_{mlp}$}
		\psfrag{f9}{$\varphi_{sh}$}
		\psfrag{f10}{$\phi_{\V w_p}$}
		\psfrag{F1}{$\V F_{\V p}$}
		\psfrag{F2}{$\V F_{\V d}$}
		\psfrag{F3}{$\V F$}
		\psfrag{F4}{$\V F_{\V p,1}$}
		\psfrag{F5}{$\V F_{\V p,2}$}
		\psfrag{F6}{$\V w_p$}
		\psfrag{F7}{$\V F_{sh}$}
		\psfrag{F8}{$s$}
		\psfrag{c1}{$\V c$}
		\psfrag{c2}{$\V c$}
		\psfrag{T1}{\footnotesize\textbf{Observation}}
		\psfrag{T2}{\footnotesize\textbf{Latent}}
		\psfrag{T3}{\footnotesize\textbf{Result}}
		\psfrag{l}{$($}
		\psfrag{r}{$)$}
}
\caption{Patterns of models designed for Neural Light Field. (a) is most widely used. 
	Our (b) learn one sphere for each $\V p$ and use a $\phi_{\V w_p}$ to map the 1D value on direction $\V d$ to 3D RGB.}
\label{fig:NSLF}
\vspace{-.6cm}
\end{figure}

Unlike previous encoding schemes that concatenate position and direction encoding, in \cref{fig:NSLF}, we propose to learn Spherical Harmonics (SH) parameters from the position encodings.
On the decoding side, we run deterministic formulas with a known SH basis.

To achieve the speedup from current techniques, we use the novel Multiresolution Hash Encoding~\cite{muller2022instant} to bear the main burden of position encoding:
$\V F_{\V p}=\phi_{p,\theta_{HG}}(\V p)$.
Such an encoding produces an encoding by interpolation of voxel features, which mitigates the global effect problem of MLPs in DSLF.
In addition, the surfaces occupy a small space in a cubic region, and thus the allocation during use is more efficient in space.
Considering that scene sequences only cover very limited direction space on surface surface, training on specific positions and directions easily leads to an arbitrary result in unseen view.
To prevent arbitrary guessing in the unseen direction, we introduce learning of SH parameters that affect the full direction.
We add an MLP-layer to generate the SH parameters: $\V F_{sh} = \phi_{mlp}(\phi_{\theta_{HG}}(\V p))=(v_\ell ^m)_{\ell: 0\leq \ell \leq \ell_{max}}^{m:\ell \leq m\leq \ell}$.
So here we are creating one sphere of latent to continuously represent a small space of $1$ dimensional color range for corresponding lights on a point. 
Then, with the known spherical harmonics function $Y_\ell ^m: \V S^2\rightarrow \mathbb C$, we extract the latent in a given direction of the sphere
utilizing the SH formula 
$\sum_{\ell =0}^{\ell _{max}}\sum_{m=-\ell }^{\ell } v^{m}_\ell  Y_\ell ^m(\V d)$ to deterministically decode on direction $\V d$:  $ s = \V F_{sh}^{T} \varphi_{sh}(\V d)$.
Meanwhile, the other branch of the encoder computes features $\V w_p$ to parameterize MLPs $\phi _{\V w_p}$. It maps (or extracts) the color range value $s$ to the color $\V c$.

Hash encoding requires normalizing the data into a unit cube~\cite{muller2022instant}. 
This means that we need to know the scale of the data.
Thus, using this model in each region avoids this problem as the regional scale is predefined.
In the following, we present a framework that operates on the growing scenes.

\subsection{NSLF with Region-wise Neural Agents}

DSLF~\cite{chen2018deep} uses MLPs to learn an entire scene.
But such a model is not robust for the incrementally creation of large scenes, because newly fed data still changes the scene globally even the non-effected parts. 
Some ideas in NeRF provide possible solutions:
(1) using voxels of multiple MLPs (KiloNeRF~\cite{reiser2021kilonerf}) or
(2) using a voxel multi-resolution representation (Instant-NGP~\cite{muller2022instant}).

\subsubsection{Multiple Asynchronous Neural Agents}
\begin{figure}[b!]
	\vspace{-.6cm}
	\centering
	\psfragfig[width=1.\linewidth]{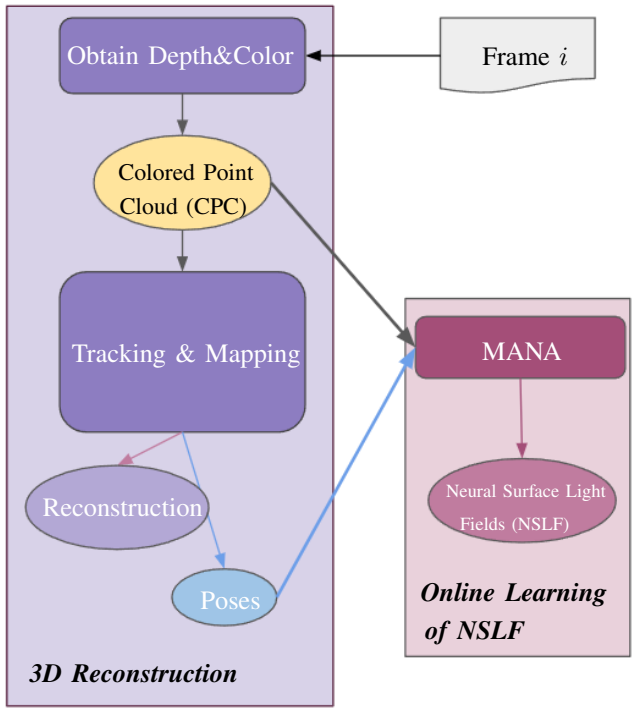}{
		\psfrag{A}{{\color{white} Obtain Depth\&Color}}
		\psfrag{B}{\small Colored Point}
		\psfrag{B1}{\small Cloud (CPC)}
		\psfrag{C}{{\color{white}Tracking \& Mapping}}
		\psfrag{D}{{\color{white}Reconstruction}}
		\psfrag{E}{{\color{white}Poses}}
		\psfrag{F}{Frame $i$}
		\psfrag{G}{{\color{white}MANA}}
		\psfrag{H}{{\scriptsize\color{white}Neural Surface Light}}
		\psfrag{H1}{\scriptsize{\color{white} Fields (NSLF)}}
		\psfrag{I}{\normalsize \textbf{\textit {3D Reconstruction}}}
		\psfrag{J}{\normalsize \textbf{\textit {Online Learning}}}
		\psfrag{J1}{\normalsize\textbf{ \textit{of NSLF}}}
	}
	\caption{MANA learns online a NSLF by serving as an \textbf{external function} to 3D reconstruction.}
	\label{fig:workflow}
\end{figure}

We use RGBD and rasterization to directly approach points on the surface during training and testing respectively. 

To speed up training while making it scalable, we divide the space equally into regions and dynamically assign models for each newly touched region. 
Each region is assigned with an Intelligent Agent (IA) that maintains its own \textbf{thread}, \textbf{neural model} and \textbf{optimizer} and is trained autonomously.
In training mode, when data is fed into an agent, it will train continuously until it reaches max-iterations.
In evaluation mode, the IA will predict input data and output colors. 
Since the implementation is done with neural network,
we will refer to such an IA as a Neural Agent (NA) in the rest of this paper. 

In addition, 
we assign an optimizer to each NA, making it an independent model that does the training itself.
Each region maintains its own neural model and optimizer to train independently.
With this feature, our model distinguishes itself from KiloNeRF which synchronizes voxels . 
And thus, our model is capable of asynchronous training.
We call it Multiple Asynchronous Neural Agents (MANA) with each NA $\V{U}=(\phi_\theta,Optim)$.

Then following KiloNeRF, we use an axis-aligned bounding box (AABB) to enclose the scene. We preset $\V b_{min}$ and $\V b_{max}$ as the minimum and maximum bound of AABB and discretize the space uniformly with the resolution $\V r=(r_x,r_y,r_z)$. We assign to each grid cell the index $\V i=(i_x,i_y,i_z)$ for the NA $\V{U}_{\V i}$. For growing scenes, we define extremely large box that cost almost nothing.

Given a RGBD frame, we unproject it to 3D space as a point cloud $\V P\in \mathbb{R}^{N_{\V P_j}\times 3}$ and their corresponding color $\V C\in \mathbb{R}^{N_{\V P}\times 3}$. The corresponding direction $\V D\in \mathbb{R}^{N_{\V P}\times 3}$ is also obtained for light fields.
Each point $\V p \in \V P$ is assigned to the corresponding region with 
\begin{equation}
v(\V p) = \lfloor (\V p - \V b_{min}) / (\V b_{max}-\V b_{min}) / \V r \rfloor.
\end{equation}
Then the color is predicted with the point $\V p_j \in \V P$ and its direction vector $\V d_j \in \V D$ as input:
\begin{equation}
\V c_{j,pred} = \phi_{\theta(v(\V p_j))}(\V p_j, \V d_j)
\label{eq:pred}
\end{equation}
Thus, for each agent $\V i$, it optimizes 
\begin{equation}
\min_{\theta(\V i)} \sum_{j,\ v(\V p_j)=\V i} || \V c_j -\V c_{j,pred}   ||^2_2
\end{equation}
independently with $\text{Optim}_{\V i}$.
\vspace{-.3cm}
\subsection{Online Learning of Surface Light Fields}
As indicated in \cref{fig:workflow}, our MANA works aside from the real-time 3D reconstruction of the model.
Note that, our model uses colored point clouds and poses sequentially from the reconstruction. 
But the training of MANA does not depend on the reconstruction result. 
The reconstruction mesh is only used during the testing, i.e., view rendering, where the surface is needed to determine the intersection points of rays.
In our implementation, the main thread is used to feed data into MANA and to perform the reconstruction.

\begin{figure}[t]
	\centering
	\psfragfig[width=1\linewidth]{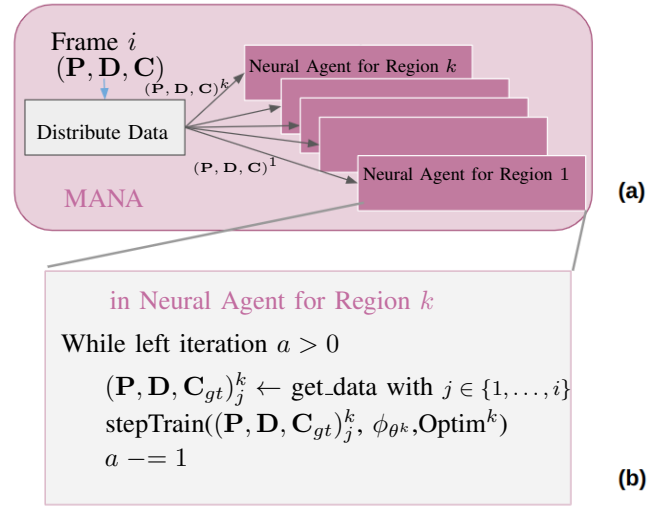}{
		\psfrag{F}{Frame $i$}
		\psfrag{p0}{$(\V P, \V D, \V C)$}
		\psfrag{pk}{\tiny{$(\V P, \V D, \V C)^{k}$}}
		\psfrag{p1}{\tiny{$(\V P, \V D, \V C)^{1}$}}
		\psfrag{NA}{{\color{magenta1}in Neural Agent for Region $k$}}
		\psfrag{l1}{While left iteration $a>0$}
		\psfrag{l2}{$(\V P, \V D, \V C_{gt})_j^k\leftarrow$ get\_data with \footnotesize{$j\in\{1,\ldots,i\}$} }
		\psfrag{l3}{stepTrain($(\V P, \V D, \V C_{gt})_j^k$, $\phi_{\theta^k}$,Optim$^k$)}
		\psfrag{l4}{$a\mathrel{-}=1$}
		\psfrag{D}{\footnotesize Distribute Data}
		\psfrag{MN}{{\color{magenta1}MANA}}
		\psfrag{nk}{\scriptsize Neural Agent for Region $k$}
		\psfrag{n1}{\scriptsize Neural Agent for Region $1$}	
	}
	\caption{Online Learning of NSLF. 
		(a) shows MANA distributing data into different Neural Agents by region. Each Neural agent maintains its thread and optimizes individually as (b).}
	\vspace{-.6cm}
	\label{fig:pipeline}
\end{figure}

We plot the diagram of MANA as a frame is fed into the pink window of \cref{fig:pipeline}.
\subsubsection{Distribution Module}

When a frame arrives, we generate a colored point cloud with direction $(\V P,\V C, \V D)$.
The distribution module feeds the points to their corresponding regions with their position. 
Together with the feed data, to ensure equal training of each region, our distribution module assigns iterations to less trained models, while setting zero iterations to more trained models. 

Since the distribution module and agents run in different threads,
the data distribution is sufficient for real-time data streaming while the training of the color models is optimized independently in the background.

\subsubsection{Asynchronous Neural Agents}

When training data of region $k$ is passed to agent $k$, it is appended to the data memory stack.
Each agent maintains its own thread to independently train the color field model $\theta$ with memory data.
As described in~\cref{fig:pipeline} (b), a thread of NA maintains the optimization of the neural model.
The newly distributed data for an agent is appended into its own memory stack. Meanwhile, left iterations {$a$} is modified for more iterations (in the experiments $a$ is set large enough such that it does not block the whole algorithm).
Then a signal is given to continue training.

\subsection{View Rendering}

After a complete online pass of the data stream, a learned NSLF and a surface are obtained.

By rasterizing with a surface mesh and camera pose as input, we obtain from the unprojected surface point $\V P$.
Rendering is then implemented by assigning points to different agents and synchronously predicting using~\cref{eq:pred}.

We show in the experiments, that in addition to the real-time training, the rendering is also done in real-time.

%% file: tex/exp.tex
\section{Results and Experiments}

Our experiments are on the scenario of \textbf{Real-time Incremental} sequences, which run the online learning modules (MANA) aside from reconstructing.

\subsection{Dataset}
\paragraph{ICL-NUIM}
ICL-NUIM~\cite{handa2014benchmark} is one of the most widely used RGBD datasets for SLAM and reconstruction purposes.
ICL-NUIM contains two synthetic scenes, i.e., room and office. 

\paragraph{Replica}
The Replica dataset~\cite{straub2019replica} provides synthetic indoor space reconstructions that contain clean dense geometry, high resolution, and dynamic textures.
It is suitable for our surface light field purpose.

\subsection{Baselines}

Our comparisons are mainly on \emph{online} learning of Neural Surface Light Fields given incremental reconstruction frames in real-time.
The reconstruction model that we work aside is a current SOTA, a large-scale real-time incremental reconstruction model, DiFusion~\cite{huang2021di}, which is based on Neural Implicit Maps.

First, we compare \emph{NSLF models} under our online learning framework (MANA) to demonstrate the advances of our proposed model.
The baselines are the recent Deep Surface Light Field models DSLF~\cite{chen2018deep}, AFFM~\cite{yu2022anisotropic}, and HashGrid (HG), an SLF model migrated from NeRF based on Instant-NGP~\cite{muller2022instant}.

Then we evaluate the model and framework \emph{as a whole} to compare the photometric performance with the latest incremental neural implicit reconstruction SOTA, NICE-SLAM~\cite{zhu2022nice}.

\input{tex/icl}

\subsection{Implementation Details}
\begin{table} [b]
\vspace{-.5cm}
  \caption {PSNR comparison on Replica sequences.} 
	\begin{adjustwidth}{0pt}{0pt}  
		\centering
		{\scriptsize
			\begin{tabular}{l|cccccccc}
				\hline
				& Ofc0
				& Ofc1
				& Ofc2
				& Ofc3
				& Ofc4
				& Rm0
				& Rm1
				& Rm2\\ \hline
				HG & 38.12&37.63&28.38&26.64&35.02&30.06	&33.51&34.37 \\
				\hline
				Ours 
				& 38.13
				& 37.85
				& 28.57
				& 26.89
				& 35.17
				& 30.20
				& 33.47
				& 34.35\\
				\hline
			\end{tabular} 
		}
		\label{tab:comp_rep1}
	\end{adjustwidth}
	
\end{table}
\begin{table}[b!]
	\vspace{-.5cm}
	\centering
	\setlength{\tabcolsep}{0.1em}
	\renewcommand{\arraystretch}{.01}
	\begin{tabular}{c |c}
		\hline
		{\large HG} &{\large \textbf{Ours}} \\\hline
		\includegraphics[width=.5\linewidth]{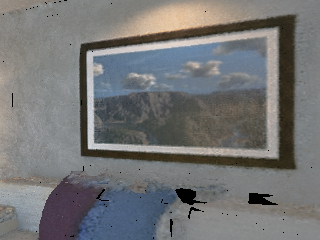}
		&\includegraphics[width=.5\linewidth]{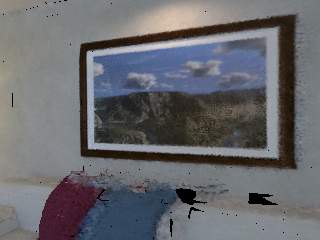}\\\hline\hline
		
		\includegraphics[width=.5\linewidth]{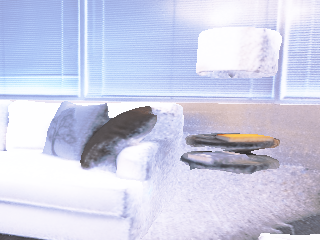}
		&\includegraphics[width=.5\linewidth]{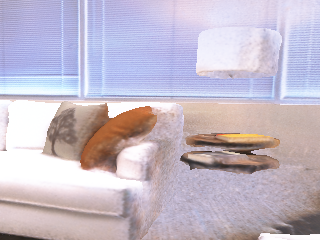}\\	\hline
		
		\includegraphics[width=.5\linewidth]{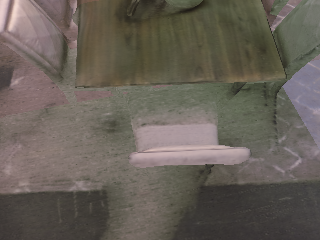}
		&\includegraphics[width=.5\linewidth]{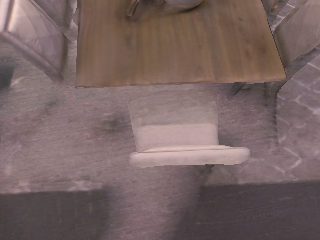}\\	\hline\hline		
		
		\includegraphics[width=.5\linewidth]{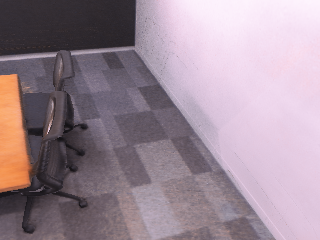}
		&\includegraphics[width=.5\linewidth]{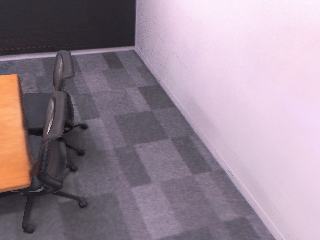}\\
		\hline
	\end{tabular}
	\captionof{figure}{Demonstration on \textbf{unseen direction} of \textbf{observed surface}.}
	\label{fig:exp:icl_unseen}
\end{table}

The implementations of DSLF and AFFM are taken from \cite{yu2022anisotropic}.
For the HG baseline, we use a multi-resolution hash grid with \textsl{resolution}=512 to encode position, and spherical harmonics to encode direction~\cite{muller2022instant}.
The decoder operates on the concatenated feature using 4 layers of MLPs with \textsl{net-width}=32. The sigmoid at the end is used to normalize value in $[0,1]$ as normalized color.
Our model is also implemented with the same multi-resolution hash grid settings. Instead of encoding directions, we learn spherical harmonics parameters and extract a value directly in that direction.
Then, the extracted latent is operated on another MLPs with learned parameters at that position for color extraction.
For all models, learning rates are set to $lr=1e-3$ to train with the Adam optimizer. 

For online learning of growing scenes, we set the region scale to $4m$.
Our MAMA works alongside the recent 3D reconstruction models (DiFusion~\cite{huang2021di}) for reconstruction support.
The main thread processes the data and passes it to reconstruction and MANA.
In the MANA, the \texttt{data\_feeding} function in the main thread takes \unit[0.01]{s} for each frame.
The agents run asynchronously in their own threading.
On the sequence, we skip and take every 20th frame into DiFusion and MANA.
During the evaluation, we infer all frames.

For comparison, we use peak signal-to-noise ratio (PSNR), structural similarity index measure (SSIM), and Learned Perceptual Image Patch Similarity (LPIPS) to evaluate the produced image quality. 

Experiments run on a computer with an AMD Ryzen 9 5950X 16-Core Processor CPU and a GTX3090Ti GPU.

\input{tex/surface_plus_light.tex}

\subsection{Online NSLFs alongside Real-time 3D Reconstruction}
\label{sec:test:online}
We implement MANA alongside Real-time 3D reconstruction for Online NSLF.
We first run MANA and the reconstruction on the real-time sequence ICL-NUIM \texttt{lrkt0n}.
For a fair comparison, we should use the same mesh for all four models.
Since ICL-NUIM does not provide a ground-truth mesh, we pre-run DiFusion to obtain the same mesh for all four NSLF models.
Four NSLF models are embedded respectively into MANA.
Then, we sequentially feed those frames into MANA.
The real-time learning skips every 20 frames and the main thread sleeps \unit[1]{s} when every 20th frame arrives.
The evaluation is done on all frames. 

\cref{tab:comp_icl} shows the photometric quantitative evaluation. 
DSLF and AFFM perform much worse than HG and ours. 
Our model shows the best performance on all metrics.
We also find this in \cref{fig:exp:icl} that
both DSLF and AFFM show blur or incorrect image prediction. While
HG and ours show a more realistic quality on the painting and sofa.

We find that HG has results, very similar to ours.
Therefore, we select HG and our model for further testing and investigation on the Replica dataset.
For the experiment on Replica which provides a ground truth mesh, to mitigate other effects, we use the ground truth mesh for prediction.
Similarly, MANA learns in real-time the sequences that skip every 20 frames and sleep \unit[1]{s} then.
In \cref{tab:comp_rep1}, HG and ours also provide very similar quality.

However, \cref{tab:comp_icl} and \cref{tab:comp_rep1} are compared on the whole sequence, which is very close to the selected trained frames.
To better demonstrate the advantage of our model, we show in \cref{fig:exp:icl_unseen} the unseen direction of the observed surface. 
For HG, these surfaces show the correct color when we capture them in the trained direction.
But we find that HG gives arbitrary results for the unseen direction of the observed surface.
For example, in the first row, HG lost the color of the painting and the sofa cushion.
In the second and third rows, HG shows arbitrary colors on the sofa and desk.
In the fourth row, HG incorrectly predicts the color of the carpet.
This problem happens to HG because the novel directions on those points are not learned.
While our model works fine.
Our model learns a sphere on each point instead of just for the observed directions.

To better support this, we further use object Chicken data from DSLF~\cite{chen2018deep}, which gives a dense view of the object.
We perform the experiment on the Chicken test set (200 frames). 
We train both HG and ours using a single frame (135th) for $1000$ iterations and render on all. 
The rendered video is demonstrated in our project page\footnote{\url{https://jarrome.github.io/NSLF-OL}}. Where HG shows high color bias when viewing angle changes while our model gives a better prediction.

From the video, most of the surface in the data is not trained. This means that quantitative evaluation of the whole sequence can hardly find the result differences.
Therefore, we compute the angle between the view directions of the inferences frame and the trained frame ($135$th). Then we threshold the angle to count only the frame result in a certain range.
In \cref{tab:chicken}, we use three options $\le 15^{\circ}$, $\le 30^{\circ}$, $\le 60^{\circ}$.
Where we find ours are always better. In addition, the large angle causes inference on the unlearned surfaces and will give a more similar score. Which is revealed by the smaller gap in the larger angle threshold.
\begin{table} [htbp]
	\caption {PSNR comparison on object sequences.}
	\centering
	\begin{tabular}{l|ccc}
		\hline
		& $\le 15^{\circ}$ & $\le 30^{\circ}$& $\le 60^{\circ}$ \\\hline
		HG&26.97&22.87&21.45\\
		Ours&27.63&23.13&21.55\\
		\hline
	\end{tabular}
	\label{tab:chicken}
\end{table}

\subsection{Comparing with Incremental Surface \& Color Reconstruction}
In the task of \emph{real-time incremental reconstruction}, surface and color are not necessarily decoupled.
For example, a recent Neural Implicit Reconstruction SOTA, NICE-SLAM provides high-quality surface and color reconstruction at the same time as online reconstruction. 
Therefore, we compare it to the \emph{SOTA of real-time incremental reconstruction} to demonstrate the advantages of our setting.
\cref{tab:comp_rep} shows the quantitative evaluation; we find that NICE-SLAM shows close performance to DiFusion combined with our MANA.
However, we find in \cref{fig:exp:fullmodel} that this is not the case. 
Our photometric result is more realistic compared to NICE-SLAM. 
The reason for this phenomenon is that DiFusion's reconstruction quality is worse than NICE-SLAM, MANA score is weakened.
For example, in the first two rows, DiFusion+MANA shows a clear plot on the wall painting while NICE-SLAM shows a very blurry result.
Also in the third and fourth rows, MANA shows fine texture on the quilt and carpet, while NICE-SLAM fails to find the detail.
However, DiFusion produces a worse reconstruction (shown in the edge of the tables and chairs), which causes defects to the NSLF rendering.

To better demonstrate the advantage of our work along the reconstruction, we use the same learned NSLFs to infer on NICE-SLAM's mesh.
This shows a better result in most of the scenes of \cref{tab:comp_rep}. The specific detail is also shown in the colored region of \cref{fig:exp:fullmodel}. 
\vspace{-.2cm}
\subsection{Time Efficiency}
Real-time reconstruction while online learning NSLF on ICL-NUIM \texttt{lrkt0n} (1508 frames) sequence in~\cref{sec:test:online} takes about \unit[28]{s}.
This means that our online learning framework incrementally trains this fixed amount of time.
Rendering a 480$\times$640 image in ICL-NUIM data takes on average \unit[0.07]{s}. 
Therefore, our model easily achieves both real-time learning (training) and rendering (inference) in a similar size.

%% file: tex/icl.tex
%

\begin{table*}[]
	\centering
	\setlength{\tabcolsep}{0.1em}
	\renewcommand{\arraystretch}{.1}
	\begin{tabular}{c | c |c |c |c}
		\toprule
		{\large DSLF~\cite{chen2018deep}}&{ \large AFFM~\cite{yu2022anisotropic}} &
		{\large HG~\cite{muller2022instant}} &{\large \textbf{Ours}} & {\large Ground Truth} \\\hline
		\includegraphics[width=.2\linewidth]{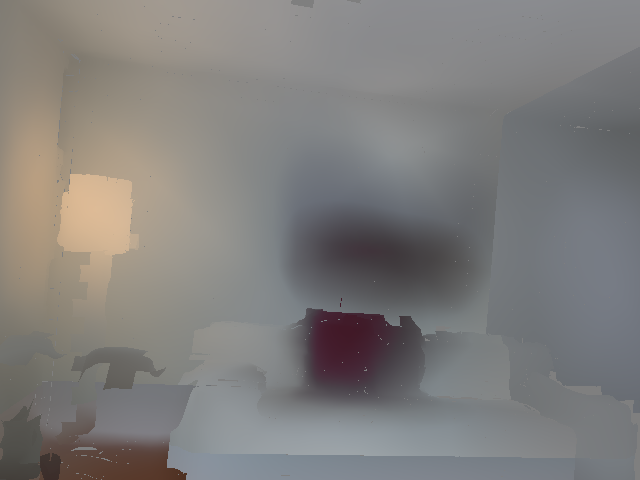}
		&
		\includegraphics[width=.2\linewidth]{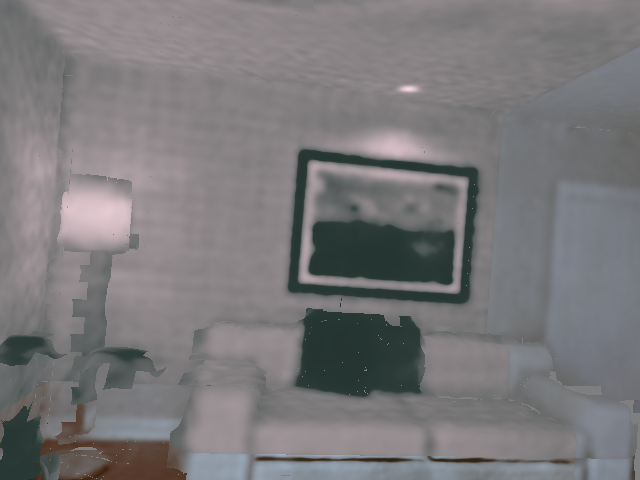}
		&\includegraphics[width=.2\linewidth]{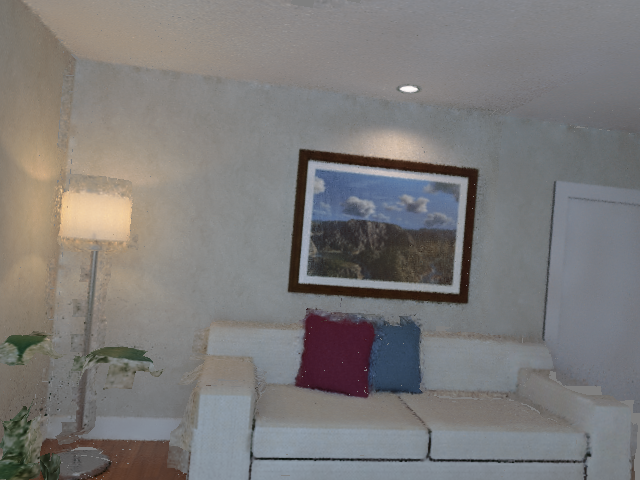}
		&
		\includegraphics[width=.2\linewidth]{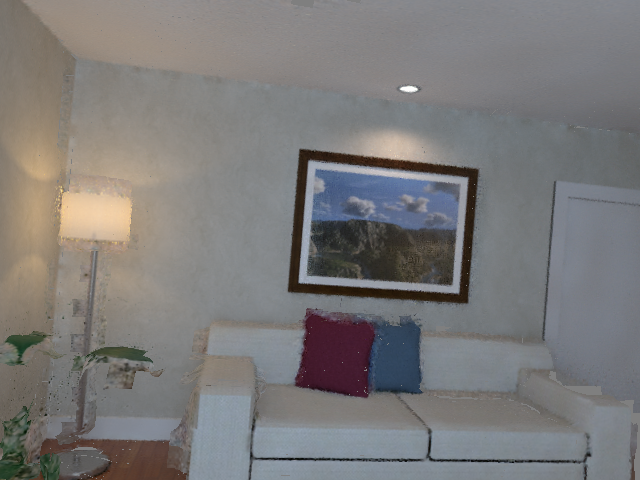}
		&\includegraphics[width=.2\linewidth]{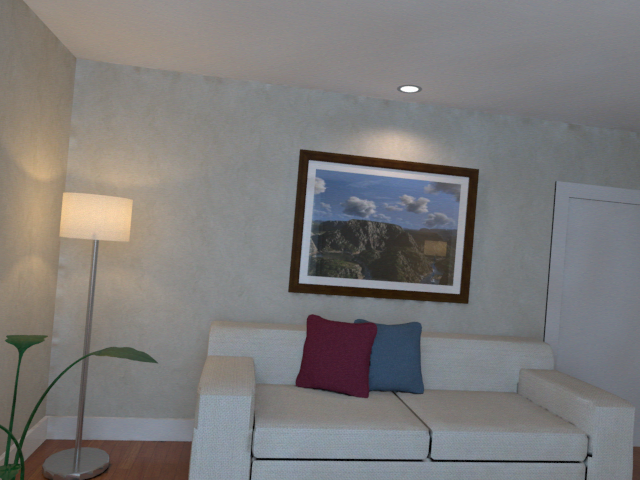}
		\\ \hline \hline
		\includegraphics[width=.2\linewidth]{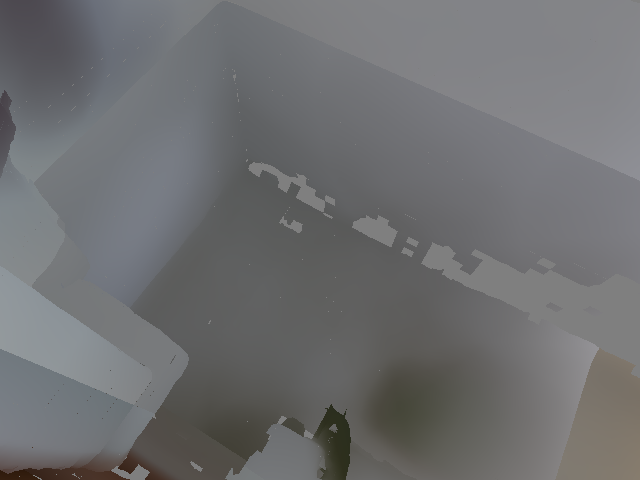}
		&
		\includegraphics[width=.2\linewidth]{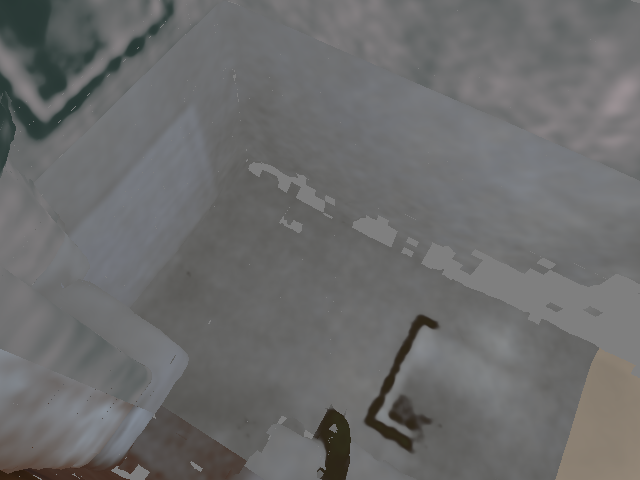}
		&\includegraphics[width=.2\linewidth]{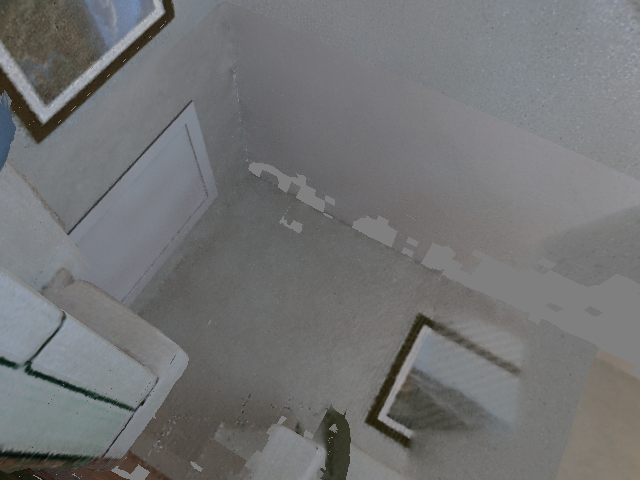}
		&
		\includegraphics[width=.2\linewidth]{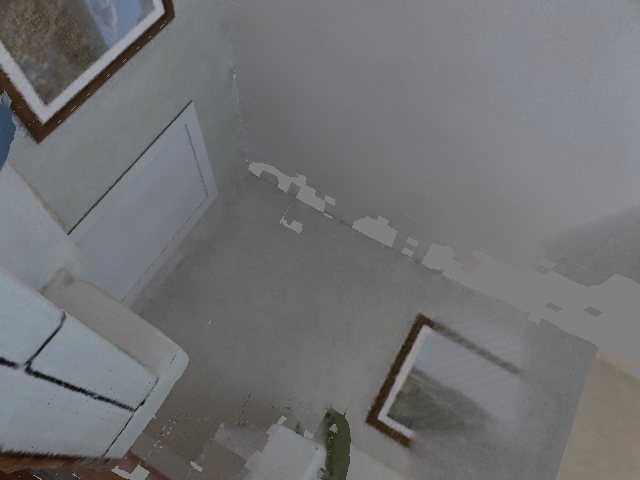}
		&\includegraphics[width=.2\linewidth]{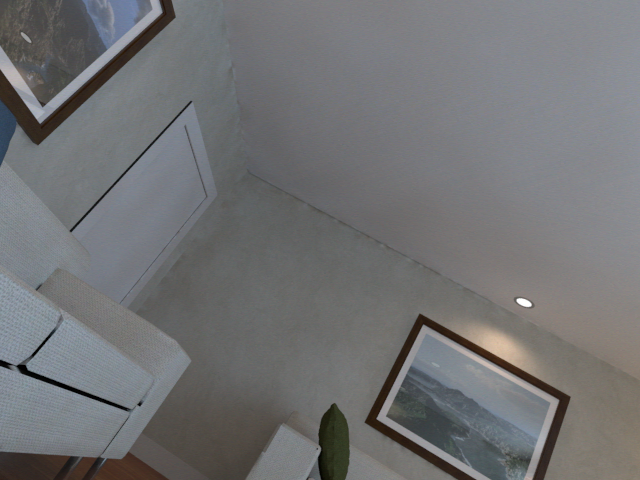}
		\\ \hline
		
	\end{tabular}
	\captionof{figure}{Demonstration on ICL-NUIM dataset lrkt0n sequence.}
	\label{fig:exp:icl}
	\vspace{-.5cm}
\end{table*}

\begin{table} [t]
  \caption {Comparison on ICL-NUIM sequence lrkt0n.} 
	\begin{adjustwidth}{0pt}{0pt}  
		\centering
		{\small
			\begin{tabular}{l|cccc}
				\hline
				& {\footnotesize DSLF~\cite{chen2018deep}}
				& {\footnotesize AFFM~\cite{yu2022anisotropic}} 
				& {\footnotesize HG~\cite{muller2022instant}}
				& {\footnotesize Ours} \\
				\hline
				PSNR $\uparrow$  &21.74 & 21.29 &28.46
				&\textbf{28.73}\\
				SSIM $\uparrow$  &0.836 & 0.846 & 0.908
				&\textbf{0.915}\\ 
				LPIPS$_{alex}$ $\downarrow$ &0.523 & 0.578 & 0.285
				&\textbf{0.277} \\
				LPIPS$_{vgg}$$ \downarrow$   &0.616 & 0.652 & 0.409
				&\textbf{0.403} \\ 
				\hline
			\end{tabular} 
		}
		\label{tab:comp_icl}
	\end{adjustwidth}
	\vspace{-.5cm}
\end{table}

%% file: tex/surface_plus_light.tex
\begin{table*}[t] 
  \caption {Full model comparison on Replica sequences. Header indicates scene names.} 
	\begin{adjustwidth}{0pt}{0pt}  
		\centering
		{\small
			\begin{tabular}{|ll|c|c|c|c|c|c|c|c|}
				\hline
				&& {Office0} & {Office1}  &{Office2} &{Office3} &{Office4} &{Room0} &{Room1} &{Room2}              \\
				
				\hline\hline
				\multirow{4}{*}{NICE-SLAM~\cite{zhu2022nice}} &
				PSNR $\uparrow$ 
				&28.38&30.68&23.90&24.88&25.18&23.46&23.97&25.94\\
				
				&SSIM $\uparrow$  &0.908&0.935&0.893&0.888&0.902&0.798&0.838&0.882 \\ 
				&LPIPS$_{alex}$ $\downarrow$ &0.386&0.278&0.330&0.287&0.326&0.443&0.401&0.315\\
				&LPIPS$_{vgg}$$ \downarrow$ &0.455&0.403&0.433&0.405&0.430&0.496&0.486&0.451
				\\
				\hline 

				\multirow{4}{*}{DiFusion\cite{huang2021di}+MANA} &
				PSNR $\uparrow$ 
				&28.59&26.70&21.10&21.89&25.74&23.24 &25.68&24.88\\
				&SSIM $\uparrow$  &0.913 &0.879&0.863&0.847&0.893&0.816&0.883&0.888 \\ 
				&LPIPS$_{alex}$ $\downarrow$ &0.371 &0.497&0.362&0.368&0.401&0.371&0.308&0.330\\
				&LPIPS$_{vgg}$$ \downarrow$ &0.395 &0.446&0.421&0.416&0.425&0.417&0.415&0.422\\				
				\hline
				
				&PSNR $\uparrow$ &30.76&30.52&22.48 &22.57 &25.94 &24.08&25.43&26.43\\

				NICE-SLAM mesh
				&SSIM $\uparrow$ & 0.942&0.927&0.891&0.872&0.911&0.819&0.879 &0.904\\

				+MANA inference
				&LPIPS$_{alex}$ $\downarrow$ &0.214&0.216&0.265&0.263&0.304&0.345&0.310&0.266\\ 

				&LPIPS$_{vgg}$$ \downarrow$ & 0.346 &0.371&0.386&0.377 & 0.406&0.416&0.420&0.403
				\\
				\hline
				
			\end{tabular} 
		}
		\label{tab:comp_rep}
	\end{adjustwidth}
\vspace{-3mm}
\end{table*}

\begin{table*}[htbp]
	\centering
	\setlength{\tabcolsep}{0.1em}
	\renewcommand{\arraystretch}{.1}
	\begin{tabular}{c |c |c| c}
	  \toprule       
		{NICE-SLAM~\cite{zhu2022nice}} &{\textbf{DiFusion\cite{huang2021di}+MANA}} & {NICE-SLAM mesh+MANA inf.} & {Ground Truth} \\[1.5ex]\hline
		
		\includegraphics[width=.24\linewidth]{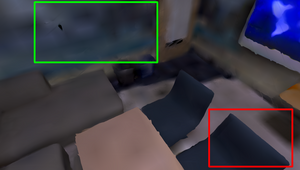}&
		\includegraphics[width=.24\linewidth]{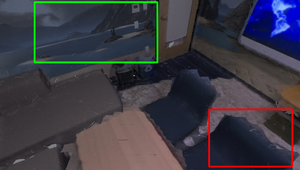}&
		\includegraphics[width=.24\linewidth]{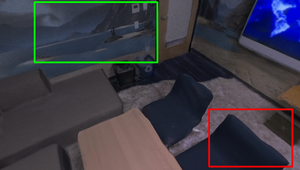}&
		\includegraphics[width=.24\linewidth]{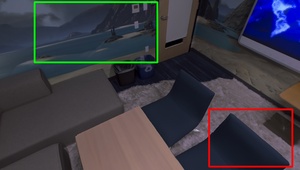}\\\hline

		\includegraphics[width=.24\linewidth]{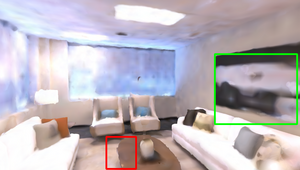}&
		\includegraphics[width=.24\linewidth]{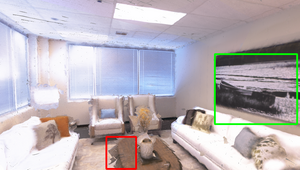}&
		\includegraphics[width=.24\linewidth]{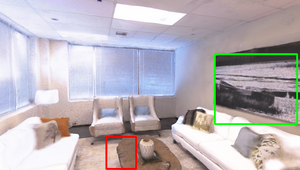}&
		\includegraphics[width=.24\linewidth]{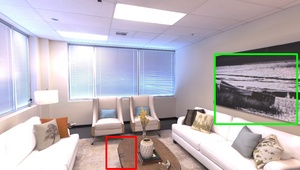}\\\hline	
		
		\includegraphics[width=.24\linewidth]{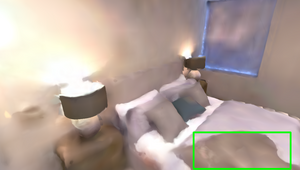}&
		\includegraphics[width=.24\linewidth]{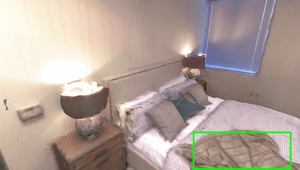}&
		\includegraphics[width=.24\linewidth]{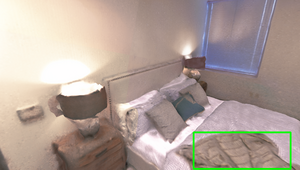}&
		\includegraphics[width=.24\linewidth]{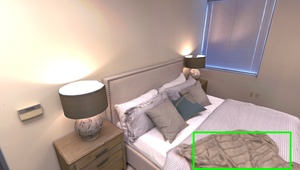}\\\hline	
		
		\includegraphics[width=.24\linewidth]{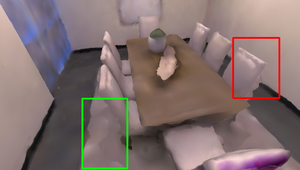}&
		\includegraphics[width=.24\linewidth]{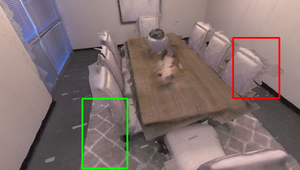}&
		\includegraphics[width=.24\linewidth]{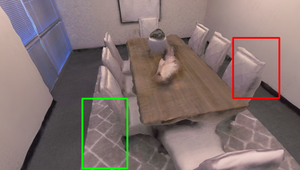}&
		\includegraphics[width=.24\linewidth]{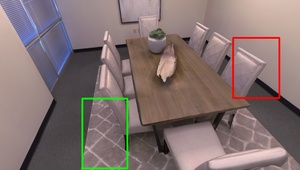}\\\hline
		 
	\end{tabular}
	\captionof{figure}{Demonstration of full model comparison. The {\color{green}green} box emphasized our better texture. With the {\color{red}red} box we shows the mesh difference between Di-Fusion and NICE-SLAM.
 	}
	\label{fig:exp:fullmodel}
	 \vspace{-5mm}
\end{table*}

%% file: v4.bbl
\begin{thebibliography}{10}
\providecommand{\url}[1]{#1}
\csname url@samestyle\endcsname
\providecommand{\newblock}{\relax}
\providecommand{\bibinfo}[2]{#2}
\providecommand{\BIBentrySTDinterwordspacing}{\spaceskip=0pt\relax}
\providecommand{\BIBentryALTinterwordstretchfactor}{4}
\providecommand{\BIBentryALTinterwordspacing}{\spaceskip=\fontdimen2\font plus
\BIBentryALTinterwordstretchfactor\fontdimen3\font minus
  \fontdimen4\font\relax}
\providecommand{\BIBforeignlanguage}[2]{{%
\expandafter\ifx\csname l@#1\endcsname\relax
\typeout{** WARNING: IEEEtran.bst: No hyphenation pattern has been}%
\typeout{** loaded for the language `#1'. Using the pattern for}%
\typeout{** the default language instead.}%
\else
\language=\csname l@#1\endcsname
\fi
#2}}
\providecommand{\BIBdecl}{\relax}
\BIBdecl

\bibitem{bylow2013real}
E.~Bylow, J.~Sturm, C.~Kerl, F.~Kahl, and D.~Cremers, ``Real-time camera
  tracking and 3d reconstruction using signed distance functions.'' in
  \emph{Robotics: Science and Systems}, vol.~2, 2013, p.~2.

\bibitem{slavcheva2016sdf}
M.~Slavcheva, W.~Kehl, N.~Navab, and S.~Ilic, ``Sdf-2-sdf: Highly accurate 3d
  object reconstruction,'' in \emph{European Conference on Computer
  Vision}.\hskip 1em plus 0.5em minus 0.4em\relax Springer, 2016, pp. 680--696.

\bibitem{yuan2022indirect}
Y.~Yuan and A.~N{\"u}chter, ``Indirect point cloud registration: Aligning
  distance fields using a pseudo third point set,'' \emph{IEEE Robotics and
  Automation Letters}, 2022.

\bibitem{niessner2013real}
M.~Nie{\ss}ner, M.~Zollh{\"o}fer, S.~Izadi, and M.~Stamminger, ``Real-time 3d
  reconstruction at scale using voxel hashing,'' \emph{ACM Transactions on
  Graphics (ToG)}, vol.~32, no.~6, pp. 1--11, 2013.

\bibitem{huang2021di}
J.~Huang, S.-S. Huang, H.~Song, and S.-M. Hu, ``Di-fusion: Online implicit 3d
  reconstruction with deep priors,'' in \emph{Proceedings of the IEEE/CVF
  Conference on Computer Vision and Pattern Recognition}, 2021, pp. 8932--8941.

\bibitem{yuan2022algorithm}
Y.~Yuan and A.~N{\"u}chter, ``An algorithm for the se (3)-transformation on
  neural implicit maps for remapping functions,'' \emph{IEEE Robotics and
  Automation Letters}, vol.~7, no.~3, pp. 7763--7770, 2022.

\bibitem{tewari2021advances}
A.~Tewari, J.~Thies, B.~Mildenhall, P.~Srinivasan, E.~Tretschk, W.~Yifan,
  C.~Lassner, V.~Sitzmann, R.~Martin-Brualla, S.~Lombardi \emph{et~al.},
  ``Advances in neural rendering,'' in \emph{Computer Graphics Forum}, vol.~41,
  no.~2.\hskip 1em plus 0.5em minus 0.4em\relax Wiley Online Library, 2022, pp.
  703--735.

\bibitem{mildenhall2020nerf}
B.~Mildenhall, P.~P. Srinivasan, M.~Tancik, J.~T. Barron, R.~Ramamoorthi, and
  R.~Ng, ``Nerf: Representing scenes as neural radiance fields for view
  synthesis,'' in \emph{European conference on computer vision}.\hskip 1em plus
  0.5em minus 0.4em\relax Springer, 2020, pp. 405--421.

\bibitem{tancik2022block}
M.~Tancik, V.~Casser, X.~Yan, S.~Pradhan, B.~Mildenhall, P.~P. Srinivasan,
  J.~T. Barron, and H.~Kretzschmar, ``Block-nerf: Scalable large scene neural
  view synthesis,'' in \emph{Proceedings of the IEEE/CVF Conference on Computer
  Vision and Pattern Recognition}, 2022, pp. 8248--8258.

\bibitem{zhang2022nerfusion}
X.~Zhang, S.~Bi, K.~Sunkavalli, H.~Su, and Z.~Xu, ``Nerfusion: Fusing radiance
  fields for large-scale scene reconstruction,'' in \emph{Proceedings of the
  IEEE/CVF Conference on Computer Vision and Pattern Recognition}, 2022, pp.
  5449--5458.

\bibitem{sucar2021imap}
E.~Sucar, S.~Liu, J.~Ortiz, and A.~J. Davison, ``imap: Implicit mapping and
  positioning in real-time,'' in \emph{Proceedings of the IEEE/CVF
  International Conference on Computer Vision}, 2021, pp. 6229--6238.

\bibitem{zhu2022nice}
Z.~Zhu, S.~Peng, V.~Larsson, W.~Xu, H.~Bao, Z.~Cui, M.~R. Oswald, and
  M.~Pollefeys, ``Nice-slam: Neural implicit scalable encoding for slam,'' in
  \emph{Proceedings of the IEEE/CVF Conference on Computer Vision and Pattern
  Recognition}, 2022, pp. 12\,786--12\,796.

\bibitem{zhang2020nerf++}
K.~Zhang, G.~Riegler, N.~Snavely, and V.~Koltun, ``Nerf++: Analyzing and
  improving neural radiance fields,'' \emph{arXiv preprint arXiv:2010.07492},
  2020.

\bibitem{wood2000surface}
D.~N. Wood, D.~I. Azuma, K.~Aldinger, B.~Curless, T.~Duchamp, D.~H. Salesin,
  and W.~Stuetzle, ``Surface light fields for 3d photography,'' in
  \emph{Proceedings of the 27th annual conference on Computer graphics and
  interactive techniques}, 2000, pp. 287--296.

\bibitem{miandji2013learning}
E.~Miandji, J.~Kronander, and J.~Unger, ``Learning based compression of surface
  light fields for real-time rendering of global illumination scenes,'' in
  \emph{SIGGRAPH Asia 2013 Technical Briefs}, 2013, pp. 1--4.

\bibitem{chen2018deep}
A.~Chen, M.~Wu, Y.~Zhang, N.~Li, J.~Lu, S.~Gao, and J.~Yu, ``Deep surface light
  fields,'' \emph{Proceedings of the ACM on Computer Graphics and Interactive
  Techniques}, vol.~1, no.~1, pp. 1--17, 2018.

\bibitem{yu2022anisotropic}
H.~Yu, A.~Chen, X.~Chen, L.~Xu, Z.~Shao, and J.~Yu, ``Anisotropic fourier
  features for neural image-based rendering and relighting,'' 2022.

\bibitem{reiser2021kilonerf}
C.~Reiser, S.~Peng, Y.~Liao, and A.~Geiger, ``Kilonerf: Speeding up neural
  radiance fields with thousands of tiny mlps,'' in \emph{Proceedings of the
  IEEE/CVF International Conference on Computer Vision}, 2021, pp.
  14\,335--14\,345.

\bibitem{liu2020neural}
L.~Liu, J.~Gu, K.~Zaw~Lin, T.-S. Chua, and C.~Theobalt, ``Neural sparse voxel
  fields,'' \emph{Advances in Neural Information Processing Systems}, vol.~33,
  pp. 15\,651--15\,663, 2020.

\bibitem{neff2021donerf}
T.~Neff, P.~Stadlbauer, M.~Parger, A.~Kurz, J.~H. Mueller, C.~R.~A. Chaitanya,
  A.~Kaplanyan, and M.~Steinberger, ``Donerf: Towards real-time rendering of
  compact neural radiance fields using depth oracle networks,'' in
  \emph{Computer Graphics Forum}, vol.~40, no.~4.\hskip 1em plus 0.5em minus
  0.4em\relax Wiley Online Library, 2021, pp. 45--59.

\bibitem{lindell2021autoint}
D.~B. Lindell, J.~N. Martel, and G.~Wetzstein, ``Autoint: Automatic integration
  for fast neural volume rendering,'' in \emph{Proceedings of the IEEE/CVF
  Conference on Computer Vision and Pattern Recognition}, 2021, pp.
  14\,556--14\,565.

\bibitem{yu2021plenoctrees}
A.~Yu, R.~Li, M.~Tancik, H.~Li, R.~Ng, and A.~Kanazawa, ``Plenoctrees for
  real-time rendering of neural radiance fields,'' in \emph{Proceedings of the
  IEEE/CVF International Conference on Computer Vision}, 2021, pp. 5752--5761.

\bibitem{hedman2021baking}
P.~Hedman, P.~P. Srinivasan, B.~Mildenhall, J.~T. Barron, and P.~Debevec,
  ``Baking neural radiance fields for real-time view synthesis,'' in
  \emph{Proceedings of the IEEE/CVF International Conference on Computer
  Vision}, 2021, pp. 5875--5884.

\bibitem{muller2022instant}
T.~M{\"u}ller, A.~Evans, C.~Schied, and A.~Keller, ``Instant neural graphics
  primitives with a multiresolution hash encoding,'' \emph{arXiv preprint
  arXiv:2201.05989}, 2022.

\bibitem{chen2022tensorf}
A.~Chen, Z.~Xu, A.~Geiger, J.~Yu, and H.~Su, ``Tensorf: Tensorial radiance
  fields,'' \emph{arXiv preprint arXiv:2203.09517}, 2022.

\bibitem{sitzmann2021light}
V.~Sitzmann, S.~Rezchikov, B.~Freeman, J.~Tenenbaum, and F.~Durand, ``Light
  field networks: Neural scene representations with single-evaluation
  rendering,'' \emph{Advances in Neural Information Processing Systems},
  vol.~34, pp. 19\,313--19\,325, 2021.

\bibitem{deng2021depth}
K.~Deng, A.~Liu, J.-Y. Zhu, and D.~Ramanan, ``Depth-supervised nerf: Fewer
  views and faster training for free,'' \emph{arXiv preprint arXiv:2107.02791},
  2021.

\bibitem{xu2022point}
Q.~Xu, Z.~Xu, J.~Philip, S.~Bi, Z.~Shu, K.~Sunkavalli, and U.~Neumann,
  ``Point-nerf: Point-based neural radiance fields,'' \emph{arXiv preprint
  arXiv:2201.08845}, 2022.

\bibitem{chen2021mvsnerf}
A.~Chen, Z.~Xu, F.~Zhao, X.~Zhang, F.~Xiang, J.~Yu, and H.~Su, ``Mvsnerf: Fast
  generalizable radiance field reconstruction from multi-view stereo,'' in
  \emph{Proceedings of the IEEE/CVF International Conference on Computer
  Vision}, 2021, pp. 14\,124--14\,133.

\bibitem{zhang2021learning}
J.~Zhang, Y.~Yao, and L.~Quan, ``Learning signed distance field for multi-view
  surface reconstruction,'' in \emph{Proceedings of the IEEE/CVF International
  Conference on Computer Vision}, 2021, pp. 6525--6534.

\bibitem{azinovic2021neural}
D.~Azinovi{\'c}, R.~Martin-Brualla, D.~B. Goldman, M.~Nie{\ss}ner, and
  J.~Thies, ``Neural rgb-d surface reconstruction,'' \emph{arXiv preprint
  arXiv:2104.04532}, 2021.

\bibitem{handa2014benchmark}
A.~Handa, T.~Whelan, J.~McDonald, and A.~J. Davison, ``A benchmark for rgb-d
  visual odometry, 3d reconstruction and slam,'' in \emph{2014 ICRA}.

\bibitem{straub2019replica}
J.~Straub, T.~Whelan, L.~Ma, Y.~Chen, E.~Wijmans, S.~Green, J.~J. Engel,
  R.~Mur-Artal, C.~Ren, S.~Verma \emph{et~al.}, ``The replica dataset: A
  digital replica of indoor spaces,'' \emph{arXiv preprint arXiv:1906.05797},
  2019.

\end{thebibliography}
